\documentclass[journal,a4paper,twoside]{IEEEtran}

\def\IEEEkeywordsname{Keywords}

\ifCLASSINFOpdf

\else
 
\fi
\usepackage[dvips]{graphicx}
\usepackage{amssymb}
\usepackage{latexsym}
\usepackage{cite}

\usepackage{graphicx}
\usepackage{amssymb}
\usepackage{amsmath}
\usepackage{amsfonts}
\usepackage{textcomp}
\usepackage{makeidx}
\usepackage{multirow}
\usepackage{verbatim}
\usepackage[usenames]{color}
\usepackage{algorithm}
\usepackage{algorithmic}

\newcommand{\CASE}[1]{\STATE \textbf{case} #1\textbf{:} \begin{ALC@g}}
\newcommand{\ENDCASE}{\end{ALC@g}}

\newcommand{\DEFAULT}{\STATE \textbf{default:} \begin{ALC@g}}
\newcommand{\ENDDEFAULT}{\end{ALC@g}}
\newcommand{\DEFAULTLINE}[1]{\STATE \textbf{default:} }
\usepackage{array}
\usepackage{eqparbox}

\begin{document}

 \begin{titlepage}
 \begin{center}
 {\Large \sc PREPRINT VERSION\\}
  \vspace{5mm}
{\huge Is swarm intelligence able to create mazes?\\}
 \vspace{10mm}
 {\Large D. Po{\l}ap, M. Wo\'{z}niak C. Napoli, E. Tramontana}\\~\\
 {\large Email: napoli@dmi.unict.it\\}~\\
 \vspace{5mm}
{\Large \sc FINAL VERSION PUBLISHED ON:\\~\\ \bf International Journal of Electronics and Telecommunications,\\ Vol. 6, n. 4, pp. 305--310 (2015)}
 \end{center}
 \vspace{5mm}
 {\Large \sc BIBITEX: \\}
 
@Article\{polap2015swarm,\\
  author =        \{Polap, Dawid and Wozniak, Marcin and Napoli, Christian and Tramontana, Emiliano\},\\
  title =         \{Is swarm intelligence able to create mazes?\},\\
  journal =       \{International Journal of Electronics and Telecommunications\},\\
  year =          \{2015\},\\
  volume =        \{61\},\\
  number =        \{4\},\\
  pages =         \{305--310\},\\
  doi =           \{10.1515/eletel-2015-0039\},\\
  url =           \{http://ijet.ise.pw.edu.pl/index.php/ijet/article/view/10.1515-elete2015-0039\}\\
\}\\

 \vspace{5mm}
 \begin{center}
Published version copyright \copyright~2015 International Journal of Electronics and Telecommunications \\
\vspace{5mm}
UPLOADED UNDER SELF-ARCHIVING POLICIES\\
NO COPYRIGHT INFRINGEMENT INTENDED \\
 \end{center}
\end{titlepage}

\title{Is swarm intelligence able to create mazes?}

\author{Dawid Po{\l}ap, Marcin Wo{\'z}niak, Christian Napoli and Emiliano Tramontana% <-this % stops a~space

\thanks{Dawid Po{\l}ap and Marcin Wo{\'z}niak are with Institute of Mathematics, Silesian University of Technology, Kaszubska 23, 44-100 Gliwice, Poland, (e-mail: Dawid.Polap@gmail.com, Marcin.Wozniak@polsl.pl)}
\thanks{C. Napoli and E. Tramontana are with Department of Mathematics and Informatics, University of Catania, Viale A. Doria 6, 95125 Catania, Italy, (e-mail: napoli@dmi.unict.it, tramontana@dmi.unict.it)}
}

\maketitle

\markboth{D. Po{\l}ap, M. Wo{\'z}niak, C. Napoli, E. Tramontana}{Is swarm intelligence able to create mazes?}

\begin{abstract}
%\boldmath
In this paper, the idea of applying Computational Intelligence in the process of creation board games, in particular mazes, is presented. For two different algorithms the proposed idea has been examined. The results of the experiments are shown and discussed to present advantages and disadvantages.
\end{abstract}


\maketitle

\section{Introduction}\label{sec:intro}
\IEEEPARstart{T}{}he first evolutionary algorithms have been shown in the 70s \cite{holland1973},\cite{baldwin1986}. Where in \cite{holland1973} J. Holland has shown that the nature is giving best optimization techniques, which are implementable in various optimization problems. Further with research on possible applications computer scientists have developed several techniques that map behavior of various animals into computer algorithms. In \cite{Yang2009} cuckoos breeding habits were implemented to search for optimum solutions, \cite{Slota2013j} and \cite{Brociek2015} present nature based algorithms applied to solve metal solidification modeling, \cite{Slota2014e} shows how to modify harmony search method to more efficiently optimize differential models of solidification models, \cite{Wozniak2015_11}, \cite{Wozniak2014_1}; \cite{NapWoa,napoli2014} and \cite{Wozniak2015_4} present evolutionary approach to model and optimize cloud based system for efficient user verification and network traffic positioning. In \cite{bartczak2012} text data clustering was optimized by application of ant colony, in \cite{bartczak2014} significant operating points were solved, while in \cite{Mandziuk2014_3} presents dedicated particle swarm modeling for dynamic routing problems. Computational intelligence based on swarm algorithms is also efficient in various image processing problems, like key-point search \cite{Wozniak2015_3}, \cite{Wozniak2014_14}, \cite{Wozniak2014_6}. Other important application of swarm intelligence leads to implementations with neural networks or other intelligent systems \cite{Wozniak2015_5}, \cite{Wozniak2015_9}, \cite{Wozniak2015_8}, \cite{MartisiusD12} and \cite{napoli2015}. Until this day these algorithms found numerous applications as an alternative to existing solutions in almost every field of science. 

One of them, where a swarm intelligence has been used, are various games. In \cite{Mandziuk2014_2} evolutionary approach was implemented to automatically solve playing strategies, while in \cite{Mandziuk2014_1} similar approach is presented to efficiently help in optimization of evaluation function for various games. Regardless of the type of game, the environment must be strongly varied in order to enhance playability. Second important feature is proper security that enables players to develop gaming playability \cite{Mandziuk2015_1}. The greater playability, the quality of the game is bigger. 

In this paper we would like present a novel approach to board games automatic development based on dedicated swarm intelligence implementation. In the case of 2D games, board is generated depending to the level of complexity. The higher the level, the board should be more difficult to pass. In most cases, each board for 2D game may be presented as a maze which is a system of many roads, where the majority does not lead anywhere. Mazes are designed in accordance with certain principles which are specified for each game separately. In general, a maze has one entrance and one exit and the player's task is to find the road leading through a maze without crossing the walls.  A similar case is for 3D games where the walls have been replaced by models of nature or some dedicated architecture models. The main problem of creating this type of board games that are focused on the playability is to reduce complexity to create an shape which may be an appropriate maze. Many maze generators are based on the graph theory which mainly uses tree search algorithms such as Prim's or Kruskal's algorithms, which are efficient for energy saving routing design \cite{Hirao2012} and gaming systems \cite{Najman2013}. In this paper, we would like to propose an alternative methods to create mazes. We base our approach on Swarm Intelligence (SI) with dedicated strategies for boards construction.  
\section{Swarm intelligence algorithms with developed procedures to compose mazes}
Swarm Intelligence is an algorithmic description of the coordinated moves that all swarm particles do together. This movement is based on communication between them, when information about surrounding conditions is passed to optimize strategy of movement. Mathematical model of this type of optimization is developed on observation of animals. Various species perform several optimization strategies to breed, feed, spread and escape. In this paper we want to concentrate on two algorithms simulating colony of ants and bees. 

In nature ants search for sources of food. Information about their locations is left for other ants in special traces of pheromones. Model of this behavior is very efficient in many optimization problems, like heat transfer modeling \cite{Slota2012u}. Similarly, it is possible to model a colony of bees. Among bees information is passed in a kind of dance that bees performs in the hive if a source of nectar was found. This algorithm can be applied i.e. for key point search in 2D pictures \cite{Wozniak2015_3}. These two algorithms can be implemented to create mazes. In the following sections we present dedicated versions developed for the purpose of maze construction.
\subsection{Artificial Ant Colony Algorithm}\label{AACAsec}
Artificial Ant Colony Algorithm (AACA) was inspired by the life of ants and more specifically the search for food. Ants are able to find a way back to the nest, mark it for other ants and also carry several times more weight of food than theirs to home. This particular modeling of nature inspired optimization AACA has been applied in various tasks, like solidification modeling \cite{Brociek2015}, image processing \cite{Wozniak2015_18} and data clustering \cite{bartczak2012}. In AACA, the movement of ants is done in random directions leaving behind a trail of pheromone which allows them to return to the nest or reconstruct a path to the food. In each case going in search of food and after finding a source of food at the way to home ant is able to move both directions because of the left pheromone track. This pheromone marking process is performed by all ants in the home. Every ant that leaves the home can follow tracks of the others. In the situation, when the number of paths is more than one, ants will choose the road in which trail pheromones is the strongest after a temporary evaporation. This guarantees that before many other ants were traveling this path, so at the and of it a large source of food can be found. Model of this marking is performed in the following iterations in the algorithm. Each iteration means that all the ants left home in the search of food and moved over the space leaving pheromone trials. In the next iteration these trials are evaluated by others and if leading to better source of food improved with new portion of pheromones or if leading to weak source abandoned. 

During the first iteration, the pheromone value is everywhere the same. In subsequent iterations, the value is updated by
\begin{equation}\label{eq:pheromone}
f^{t+1}(\mathbf{x_i},\mathbf{x_j})=(1-\rho)f^{t}(\mathbf{x_i},\mathbf{x_j})+\Gamma_i^{t},
\end{equation}
where $\rho$ is evaporation rate, $t$ is the current iteration and $\Gamma$ is th distance between the $\mathbf{x_i}$ and all individuals $n$ in the population. The distance is calculated by
\begin{equation}\label{eq:distance}
\Gamma_i^{t}=\sum_{i=1}^n\frac{1}{L_{ij}^{t}},
\end{equation}
where $L^t_{ij}$ is the length of the path between ants $i$ and $j$, which is defined as Cartesian metric 
\begin{equation}\label{eq:lengtgh}
L_{ij}=\lVert \mathbf{x_i} - \mathbf{x_j} \rVert = \sqrt{\sum^{2}_{k=1}(x_{i,k}-x_{j,k})^2},
\end{equation}
where $\mathbf{x_i}$ and $\mathbf{x_j}$ are points in $R\times R$ space, $x_{i,k}, x_{k,j}$-k-th components of the spatial coordinates $\mathbf{x_i}$ and $ \mathbf{x_j}$ representing insect.

The probability of choosing the road to the ant $\mathbf{x_j}$ by $\mathbf{x_i}$ is calculated by
\begin{equation}\label{eq:choosepath}
p^{t}(\mathbf{x_i},\mathbf{x_j})=\frac{[f^{t}(\mathbf{x_i},\mathbf{x_j})]^\alpha\left[\frac{1}{L_{ij}^{t}}\right]^\beta}{\sum_{\alpha\in N^k_i}\left([f^{t}(\mathbf{x_i},\mathbf{x_\alpha})]^\alpha\left[\frac{1}{L_{i\alpha}^{t}}\right]\right)},
\end{equation}
where $N^k_i$ is a set of unknown roads for $k$ ant which lead to the $i$, $\alpha$ means the impact of left pheromones.

The movement of ants is based on the highest probability $p^t$ and it is defined as
\begin{equation}\label{eq:move}
\mathbf{x_i^{t+1}}=\mathbf{x_i^{t}}+\text{sign}(\mathbf{x_i^{t}}(\text{ind}(t))-\mathbf{x_i^{t}}),
\end{equation}
where $\text{ind(t)}$ means a set of neighbor indicts after sort. Implementation of the algorithm is shown in Algorithm \ref{AACAAlgorithm}.
\begin{algorithm}[!ht]
\caption{AACA to create mazes}
\label{AACAAlgorithm}
\begin{algorithmic}[1]
\STATE Start,
\STATE Define all coefficients: $n$ size of workers population, $\alpha$ impact of left pheromones, $\rho$ evaporation rate, $\zeta$ the minimum value of the pheromone,$r$ number of random alleys, 
\STATE Create array with $\alpha$ values and two different exits,
\WHILE{the queen does not pass the entire maze}
	\STATE{Update pheromone values using \eqref{eq:pheromone},}
	\STATE{Calculate distances between worker ants \eqref{eq:lengtgh},}
	\STATE{Calculate possible path to follow by worker $i$ to location $j$ $p^{t}(\mathbf{x_i},\mathbf{x_j})$ using \eqref{eq:choosepath},}
	\STATE{Determine the best position to follow,}
	\STATE{Move population of workers using \eqref{eq:move},}
\ENDWHILE
\STATE Recalculate the value of pheromone according to \eqref{eq:map},
\STATE Return array of recalculated values of pheromone.
\STATE Stop.
\end{algorithmic}
\end{algorithm}

\subsection{Artificial Bee Colony Algorithm}\label{ABCAsec}
Artificial Bee Colony Algorithm (ABCA) reflects the behavior of honey bees during the search for food. When bees are looking for nectar sources, they can communicate with each other. One of the most common communication techniques is the waggle dance, during which bees inform each other about the quality of the found source, distance from the hive and direction. Waggle dance allows to find the location where the best nectar is to be found. In the algorithm, there are different types of bees because of their role in the hive. The first group are scouts who look for food at random way and communicate with others by waggle dance in the hive. The second group are an onlookers who choose the best areas for exploration after the watch of the dance. The last are employed bees who are looking for the best quality nectar in the designated areas. All of them communicate to pass the information. Model of this process is performed in the following iterations in the algorithm. Each iteration means that all the bees left home in the search of food and came back pass the information to the others. In the next iteration other bees fly to these locations to search for food. If a better source of food is found this is information is passed at the end of iteration.

In the first stage of the algorithm, we create an array that reflects the map of meadow where bees are moving in search of the best flowers. At the beginning, all fields (except entrance and exit) has the same value of $\zeta$. The exits are marked threefold value of $\zeta$. Depending on the quality of the place, each bee may modify this value by the following equation
\begin{equation}\label{mapModification}
\Theta(\zeta)=
\begin{cases}
\zeta+0,1 \quad\quad\quad\text{if}\quad\quad \Gamma(\mathbf{x_i}, \zeta)<0,5\\ 
\zeta-0,05 \quad\quad\quad\text{if}\quad\quad \Gamma(\mathbf{x_i}, \zeta)>0,5
\end{cases}
,
\end{equation}
where $f(\mathbf{x_i},\zeta)$ is called fitness function and it is calculated by
\begin{equation}\label{fitness}
\Gamma(\mathbf{x_i},\zeta)=\zeta\sqrt{(x_{r,0}-x_{i,0})^2+(x_{r,1}-x_{i,1})^2},
\end{equation}
where $r$ is a index of exit to which the distance is the smallest from current position. 

In the implementation of the algorithm, we assume that each bee is a point in space. When the new sources are found, the bees can leave the current position towards better. Bees are looking for a new source when they receive information from the hive after watching the waggle dance. At the end of each iteration, all the bees are compared in order to obtain information about the location of the best source. It is interpreted as the waggle dance. An onlooker bee choose the best area through by
\begin{equation}\label{ABCAwaggle}
p(\mathbf{x_i})=\frac{\Gamma(\mathbf{x_i},1)}{\sum_{i=1}^{i=n} \Gamma(\mathbf{x_i},1)}.
\end{equation}
After gaining information about the fitness of bees, they are sorted to find the best positions. After that, onlookers can move in this direction. The movement of bees is made by 
\begin{equation}\label{ABCAmotion}
\mathbf{x_i}^{t+1}=\mathbf{x_i}^t+\alpha_{k}\cdot \varDelta\mathbf{x_{ik}},
\end{equation}
where $k$ is a random index from the set of the best sources, $\alpha_k$ is a random number from $\langle -1,1\rangle$ and $\varDelta\mathbf{x_{ik}}$ is obtained by
\begin{equation}\label{ABCAcontact}
\varDelta\mathbf{x_{ik}}=(x_{ij}-x_{kj}),
\end{equation}
where $j$ is randomly chosen spatial coordinate of the chosen bee. The implemented algorithm is illustrated in Algorithm \ref{ABCAAlgorithm}. The algorithm returns an array that is a map on which the bees moved.
\begin{algorithm}[!ht]
\caption{ABCA to create mazes}
\label{ABCAAlgorithm}
\begin{algorithmic}[1]
\STATE Start,
\STATE Define all coefficients: $n$ size of population, $m$ - number of chosen best bees, $\zeta$ the minimum value to create a wall, $r$ number of random alleys, 
\STATE Create position array with value $\zeta$ and two different exits,
\STATE Create population,
\WHILE{the queen does not pass the entire maze}
	\STATE{Evaluate the values in array according to \eqref{mapModification}}
	\STATE{Evaluate population using \eqref{ABCAwaggle}}
	\STATE{Sort $bees$ according to the value of location,}
	\STATE{Select $m$ best locations among all $bees$,}
	\STATE{Other $bees$ replace with randomly selected bees}
	\STATE{using \eqref{ABCAmotion} move the $bees$ toward nectar source (exits) defined in \eqref{ABCAcontact}},
\ENDWHILE
\STATE Recalculate the value from position array according to \eqref{eq:map},
\STATE Return array,
\STATE Stop.
\end{algorithmic}
\end{algorithm}

\subsection{An Adaptation of AACA and ABCA to Create Mazes}
An adaptation of presented Swarm Intelligence algorithms to create mazes requires additional operations to obtain picture of the maze. Each of the presented algorithms returns an array of values (ants – an array of pheromones, bees – the map of meadow) which represent the maze. Each of these values is calculated by the following function
\begin{equation}\label{eq:map}
\Lambda(y)=
\begin{cases}
y\in\langle 0,\kappa) \hbox{\quad empty space} 
\\
y\in(\kappa,1\rangle \hbox{\quad walls }
\end{cases},
\end{equation}
where $\kappa$ is the limit value for which it will create a wall of the maze.
\begin{algorithm}[!ht]
\caption{The algorithm for maze design using Swarm Intelligence}
\label{ALLIN}
\begin{algorithmic}[1]
\STATE Start,
\STATE Use Algorithm \ref{AACAAlgorithm} or Algorithm \ref{ABCAAlgorithm} to create an array representing a maze,
\STATE Create a bitmap $I$,
\FORALL{value $v$ in array}
	\IF{$v$ is $1$}
		\IF{$r>0$}
			\IF{random value is less than $0.5$}
				\STATE Create a wall.
			\ENDIF
		\ELSE
			\STATE Create a wall.
		\ENDIF
	\ELSE
		\FORALL{neighbor of $v$}
			\IF{neighbor is $1$}
				\STATE Create a wall.
			\ENDIF
		\ENDFOR
	\ENDIF
\ENDFOR
\STATE Stop.
\end{algorithmic}
\end{algorithm}
Presented AACA and ABCA algorithms are used in Algorithm \ref{ALLIN} to create various mazes. However to improve the efficiency of board games creation we have implemented a dedicated stop construction process. This is based on stop condition that is used to verify if it is still necessary to continue to compose of the paths in the maze.
\subsection{Stop Condition}
To adapt the AACA and ABCA algorithms described in subsection \ref{AACAsec} and \ref{ABCAsec}, a dedicated stop condition must be adjusted to create a road from entrance to exit of the maze. For this purpose, the queen (a moving supervisor) is added to each of the algorithms. After each iteration, the queen is called to check whether exists a passage through constructed maze. If the passage exits, the algorithm is complete and the maze is ready. Otherwise, the next iteration begins since the queen is not satisfied with the work of their subordinates.

Seeking for an exit, the queen moves according to the Cartesian metric defined in \eqref{eq:lengtgh} where we assume that 
\begin{equation}\label{eq:StopCondition}
L_{ij}=1.
\end{equation}
The possibility of passing diagonally across the maze is locked in this way. That helps to ensure that the queen moves only in horizontal or vertical line. The whole stop condition algorithm is shown in Algorithm \ref{march}.
\begin{algorithm}[!ht]
\caption{The imperial march of the Queen}
\label{march}
\begin{algorithmic}[1]
\STATE Start,
\STATE Create an array in accordance with (\ref{eq:map}),
\STATE Find all the entrances to the maze, 
\FORALL{entry to the maze}
	\WHILE{there is no other movement}
\STATE{Find neighboring fields,}
\STATE{Remove fields in a row, in which the Queen was in the previous step,}
\FORALL{neighboring fields}
	\IF{equation (\ref{eq:StopCondition}) is not true or the field is a wall}
		\STATE{Delete field,}
	\ENDIF
\ENDFOR
	\STATE{Select at random one of the existing movements,}
	\ENDWHILE
	\IF{the last field is one of the entrances}
		\STATE{End of the algorithm - there is a way out of the maze,}
	\ENDIF
\ENDFOR
\STATE{End of the algorithm - there is no way out of the maze,}
\STATE Stop.
\end{algorithmic}
\end{algorithm}
For even more complicated constructions of mazes, the $r$ parameter can be added. Parameter $r$ will represent the number of random alleys which is the number of walls to be removed in random way. In this case, when an initial board is created we can improve efficiency of the swarm to create the passage.
\section{Experiments}
\begin{figure}[!ht]
 	\centering
 	\includegraphics[width=3.3in]{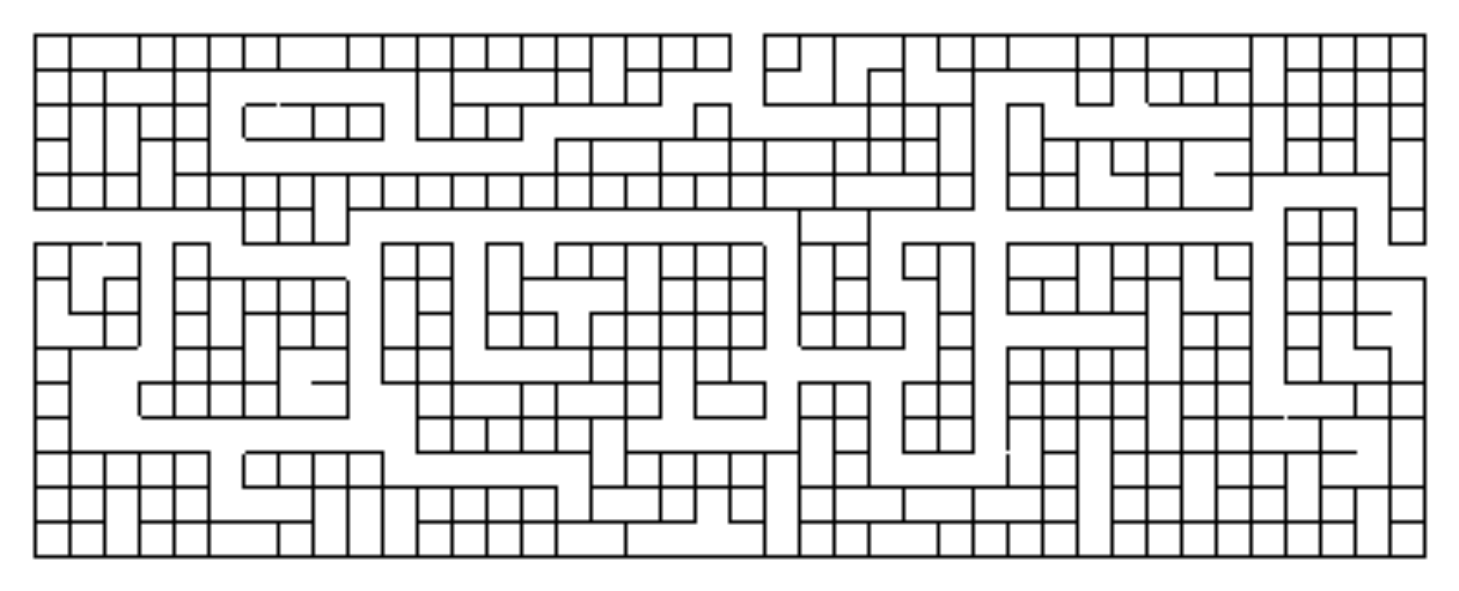}
  \caption{An example of $40\times 15$ maze generated by the AACA.}\label{fig:maze1}
\end{figure}

\begin{figure}[!ht]
 	\centering
 	\includegraphics[width=3.3in]{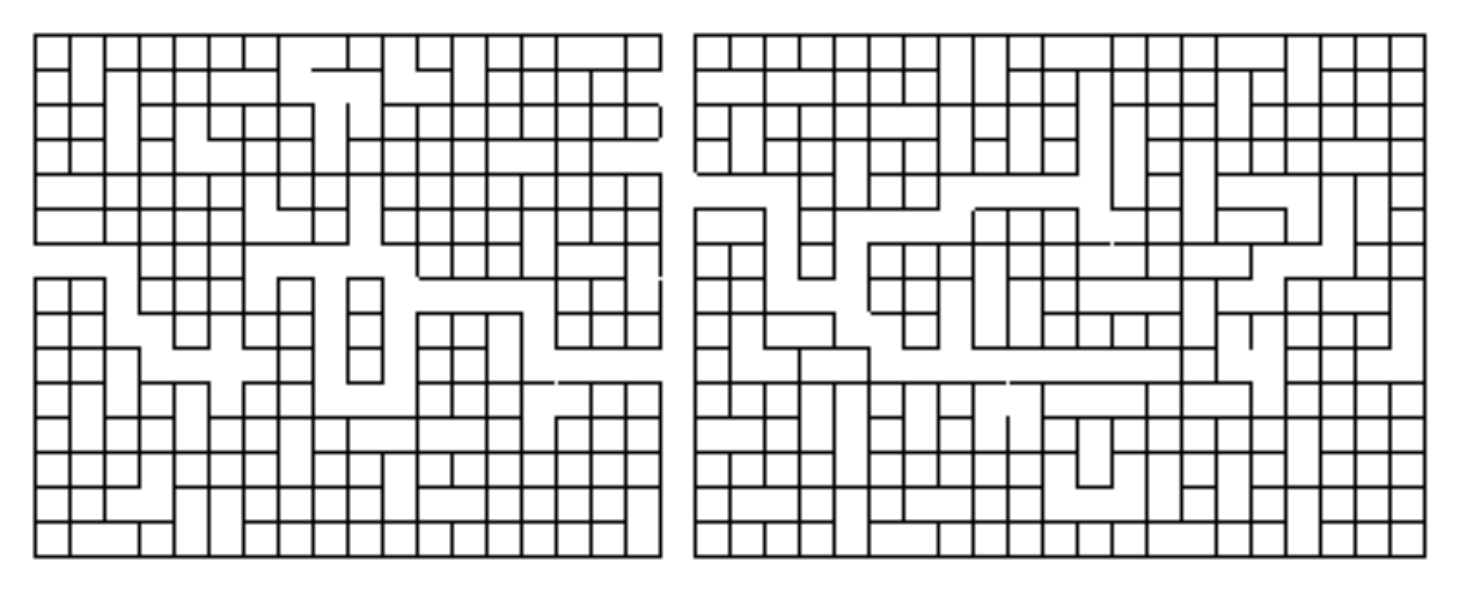}
  \caption{An example of $40\times 15$ maze generated by the ABCA.}\label{fig:maze2}
\end{figure}

\begin{figure*}[!ht]
 	\centering
 	\includegraphics[width=4.3in]{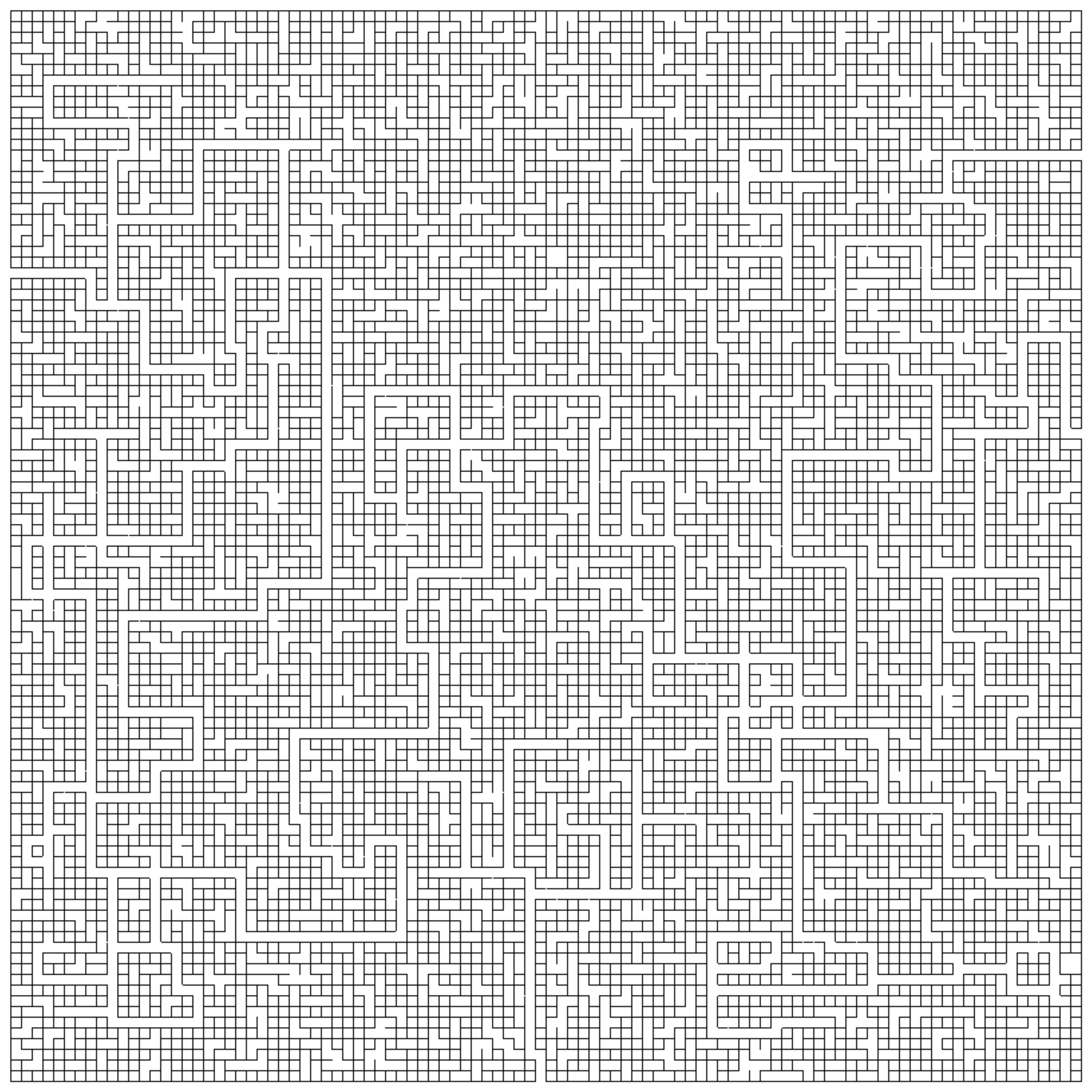}
  \caption{An example of $100\times 100$ maze generated by the AACA.}\label{fig:maze3}
\end{figure*}

\begin{figure*}[!ht]
 	\centering
 	\includegraphics[width=4.3in]{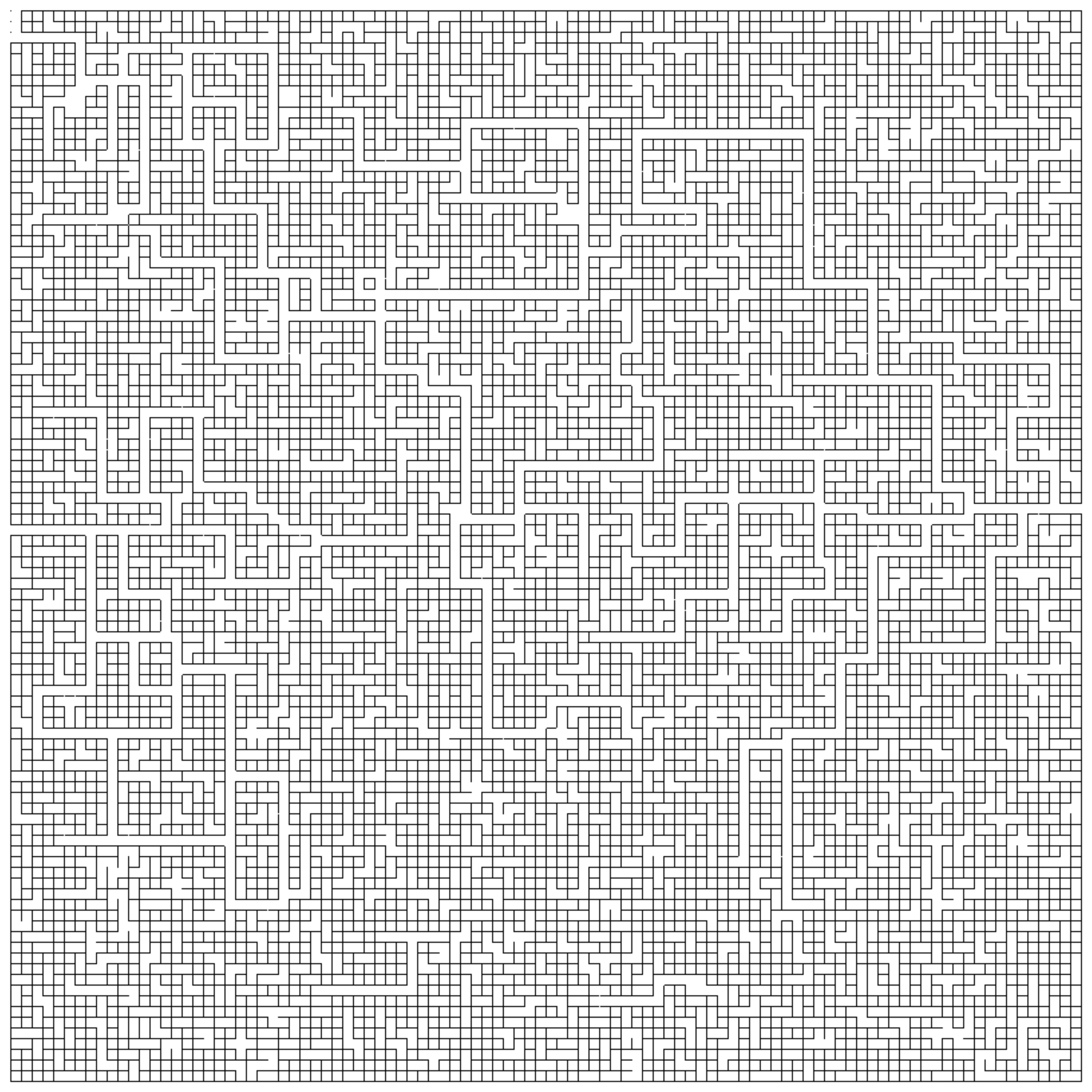}
  \caption{An example of $100\times 100$ maze generated by the ABCA.}\label{fig:maze4}
\end{figure*}

\begin{figure}[!ht]
 	\centering
 	\includegraphics[width=3.5in]{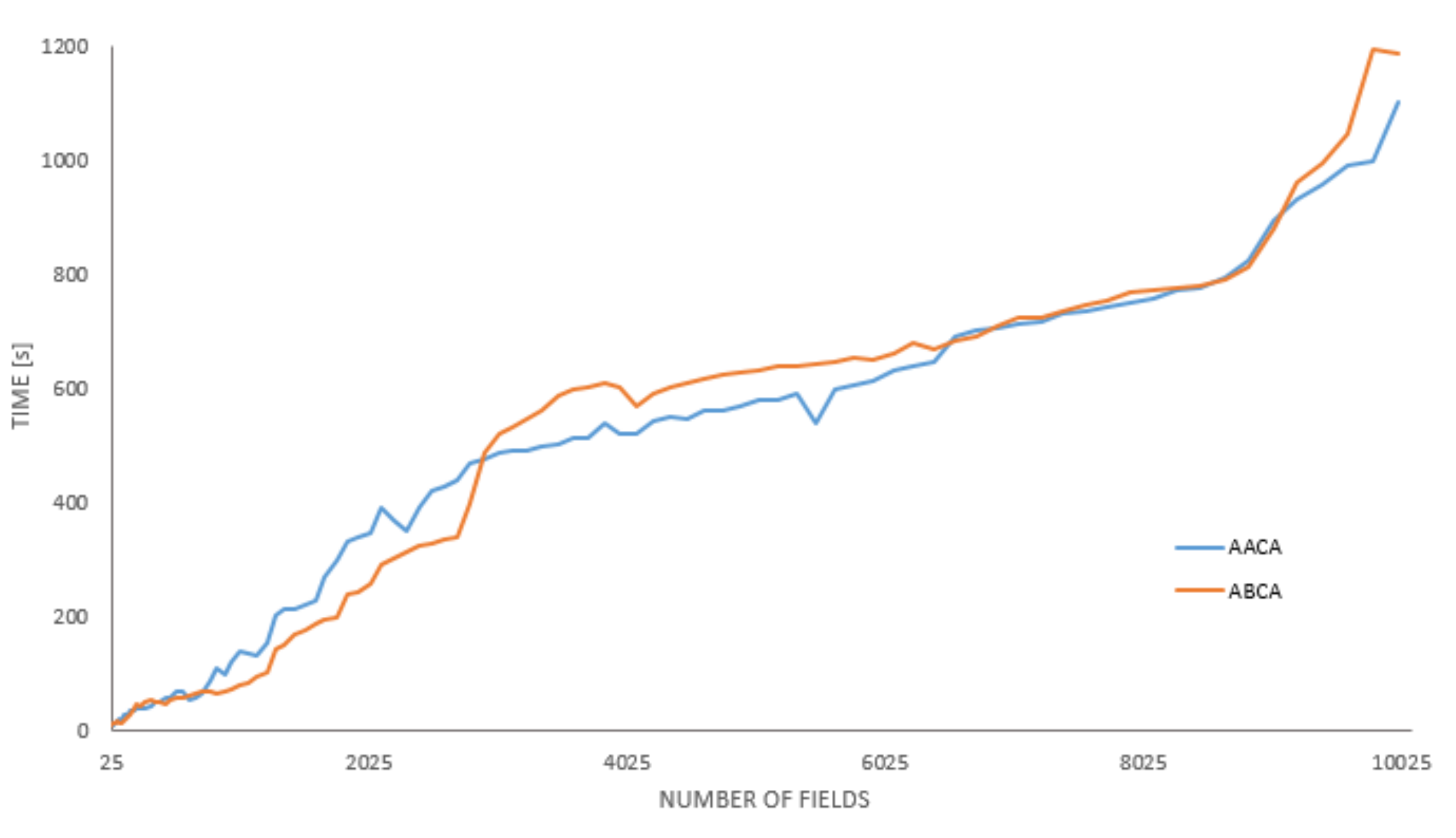}
  \caption{Time to generate the maze for applied algorithms.}\label{timeChart}
\end{figure}

Presented approach was implemented to create various mazes in different resolutions and combinations. We have performed benchmark tests on various dimensions of boards. Let us present sample results for:
\begin{itemize}
	\item square mazes (Fig. \ref{fig:maze1} and Fig. \ref{fig:maze2}) with parameters values $n=30$, $r=20$, $\zeta=$, ABCA ($m=10\%n$), AACA ($\alpha=0,4$, $\rho=0,3$),
	\item rectangle mazes (Fig. \ref{fig:maze3} and Fig. \ref{fig:maze4}) with parameters values $n=200$, $r=120$, $\zeta=$, ABCA ($m=10\%n$), AACA ($\alpha=0,4$, $\rho=0,3$).
\end{itemize}
% %
% %
% %
% %
Fig. \ref{timeChart} presents a chart of time comparison for creation of mazes by both AACA (blue line) and ABCA (orange line).
Up to 3000 fields in the maze AACA is less efficient in time comparison to ABCA. If the maze we create contains between 3000 and 6000 fields AACA take advantage. From 6000 to 9000 fields both algorithms present similar time efficiency. Above 10000 fields AACA is much faster.
% %
\section{Final remarks}
In the research, many experiments have been realized for different dimensions of mazes. The results allow to conclude that Swarm Intelligence is capable to create mazes. Fig \ref{timeChart} shows that the maze of complex up to 3000 fields is best to use ABCA presented in section \ref{ABCAsec}, but for larger mazes AACA presented in section \ref{AACAsec} seems to be a better choice. 

Generating maze is quite a complex process but the results are very satisfied – mazes are created randomly which gives a feature of uniqueness. As a result, the algorithms presented in this paper can be used to create not only mazes but environments for 2D or even 3D games and also applicable to smart cities models for prediction and evaluation purposes.

\section*{Acknowledgements}
This work has been partially supported by project PRIME funded by the Italian Ministry of University and Research within POR FESR Sicilia 2007-2013 framework.
%\IEEEtriggeratref{6}
\bibliographystyle{IEEEtran}
\bibliography{wozniak_publications_daty,gapublications,slota_publications,otherseswa,rutkowski,mandziuk,RobertasDamasevicius,christian}

\end{document}